\documentclass[conference]{IEEEtran}
\IEEEoverridecommandlockouts
\usepackage{cite}
\usepackage{mathtools}
\usepackage{amsmath,amssymb,amsfonts}
\usepackage{algorithmic}
\usepackage{graphicx}
\usepackage{hyperref}
\usepackage{textcomp}
\usepackage{xcolor}
\usepackage{fancyhdr}
\usepackage{float}
\def\BibTeX{{\rm B\kern-.05em{\sc i\kern-.025em b}\kern-.08em
    T\kern-.1667em\lower.7ex\hbox{E}\kern-.125emX}}

\begin{document}

\title{A Comprehensive Survey of Regression Based Loss Functions for Time Series Forecasting}

\author{\IEEEauthorblockN{Aryan Jadon}
\IEEEauthorblockA{\textit{San José State University} 
\\
San Jose, USA \\
aryan.jadon@sjsu.edu}

\and

\IEEEauthorblockN{Avinash Patil}
\IEEEauthorblockA{\textit{Juniper Networks Inc} \\
Sunnyvale, USA \\
patila@juniper.net}

\and

\IEEEauthorblockN{Shruti Jadon}
\IEEEauthorblockA{\textit{Juniper Networks Inc} \\
Sunnyvale, CA \\
sjadon@juniper.net}
}

\maketitle
\begin{abstract}

Time Series Forecasting has been an active area of research due to its many applications ranging from network usage prediction, resource allocation, anomaly detection, and predictive maintenance. Numerous publications published in the last five years have proposed diverse sets of objective loss functions to address cases such as biased data, long-term forecasting, multicollinear features, etc. In this paper, we have summarized 14 well-known regression loss functions commonly used for time series forecasting and listed out the circumstances where their application can aid in faster and better model convergence. We have also demonstrated how certain categories of loss functions perform well across all data sets and can be considered as a baseline objective function in circumstances where the distribution of the data is unknown. Our code is available on \href{https://github.com/aryan-jadon/Regression-Loss-Functions-in-Time-Series-Forecasting-Tensorflow}{GitHub}.

\end{abstract}


\section{Introduction}

A Time Series Forecasting approach involves making predictions using historical time-stamped data. It entails analyzing historical data, constructing models, and applying learned patterns and findings to influence future strategic decisions. In past, time series forecasting was dominated by linear methods because of their model interpretability, explainability, and efficiency for a wide range of simpler forecasting problems. However, In many of the latest applications, time series forecasting uses technologies like Machine Learning, Deep Learning, Gaussian Processes, and Artificial Neural Networks. The choice of the loss function also known as objective function is extremely important while working on large-scale time series forecasting problems as they instigate the learning process of the algorithm. Therefore, since 2012, researchers have experimented with various specific loss functions to improve results for their datasets. In this paper, we have summarized such widely used 14 loss functions that have been proven to provide state-of-the-art results in different domains.

We have also explored the conditions to determine which loss function might be helpful in a scenario and compared the performance of all loss functions on four different datasets using Temporal Fusion Transformer\cite{fusion_transformers} for forecasting. Datasets used in experiments are Electricity Load Diagrams 2011-2014 Dataset\cite{misc_electricity_load_diagrams}, which has electricity consumption data points of 370 points/clients, PEMS-SF Dataset\cite{misc_pems}, which has 15 months worth of daily data that describes the occupancy rate of different car lanes of the San Francisco bay area freeways across time, Corporación Favorita Dataset\cite{Realized72} which has Grocery Sales Data and Volatility Dataset\cite{Corporac47} which has daily non-parametric measures of how volatility financial assets or indexes were in the past. The code implementation of experiments is available at \href{https://github.com/aryan-jadon/Regression-Loss-Functions-in-Time-Series-Forecasting-Tensorflow}{https://github.com/aryan-jadon/Regression-Loss-Functions-in-Time-Series-Forecasting-Tensorflow}

In general, Time series forecasting problems can be converted and worked into a regression-type predictive modeling problem. However, the above mentioned components of a time series data make the forecasting problem more complex for simple linear methods. Therefore, an ideal model and a good objective function should be able to learn trend and seasonality of data, while avoiding random noise. With the ease of computation power, the model choice can be varied as much the complexity of data requires, but there is still some uncertainty about the choice of loss functions. In the next section, we have detailed about 14 widely used loss function for forecasting problem. We aim to help industry professionals, as well as researchers by guiding them to correct objective function/loss function which can reduce the set of experiments, converge the model faster, and overall help researchers lead their research in better direction.

The rest of the paper is outlined as follows, Section II provides explanation about time series data, and Section III describes the regression based loss functions in detail. Section IV reports the experiments and results. Section V discusses the limitations of this approach and concludes the paper.


\section{Time Series Data}
Regression modeling is a form of predictive modeling technique used to estimate the relationship between two or more variables. It is a supervised learning technique that can be defined as statistical techniques used to model the relationship between the dependent real numbered variable $y$ and the independent variables $x_i$. Time series data varies a bit different from general regression based data, as there is a temporal information added to the features making the objective a bit more complex. A Time Series data have the following components - 
\begin{enumerate}
\item Level - Every time series has a baseline. These baselines are called levels; to these levels we add different components to form a complete time series.
\item Cyclicity - Time series data also has a pattern called cyclicity which repeats aperiodically, meaning it does not occur at the same fixed interval.
\item Trend - It indicates whether a time series is increasing or decreasing over a period of time. That is, it has an upward (increasing) or downward (decreasing) trend.
\item Seasonality - Patterns that repeat over a period of time are called seasonality.
\item Noise - After extracting level, cyclicity, trend, and seasonality, what is left is noise. Noise is an entirely random variation in data.
\end{enumerate}

The fundamental goal of every classical machine learning model is to improve the model's selected metrics and reduce the loss associated with it. A vital component of a machine learning or deep learning model used for time series forecasting is the loss function, a model's performance is measured against it, and the parameters it chooses to learn will be based on the minimization of function. 

\section{Regression Loss Functions}

Regression loss functions that can be used for improving the performance evaluation in regression analysis are:

\begin{enumerate}
\item Mean Absolute Error (MAE)
\item Mean Squared Error (MSE)
\item Mean Bias Error (MBE)
\item Relative Absolute Error (RAE)
\item Relative Squared Error (RSE)
\item Mean Absolute Percentage Error (MAPE)
\item Root Mean Squared Error (RMSE)
\item Mean Squared Logarithmic Error (MSLE)
\item Root Mean Squared Logarithmic Error (RMSLE)
\item Normalized Root Mean Squared Error (NRMSE)
\item Relative Root Mean Squared Error (RRMSE)
\item Huber Loss
\item Log Cosh Loss
\item Quantile Loss

\end{enumerate}


\subsection{Mean Absolute Error (MAE)}
Absolute Error, also known as L1 loss, is the absolute difference between a predicted value and the actual value.
$$ L1 = \mid y_{actual} - y_{predicted}\mid$$
The aggregation of all loss values is called the cost function, where the cost function for Absolute Error is known as Mean Absolute Error.
$$ MAE = \frac{1}{N} \sum_{i=1}^{N}|y_i - \hat{y_i}| $$

Here, $N$ is the number of data samples, $y_i$ is the true value, and $\hat{y_i}$ is the predicted value.

Mean Absolute Error is a simple yet robust loss function used in regression models. Because of outliers, variables in regression problems may not be strictly Gaussian \cite{Qi_2020}. Refer to table \ref{table1} to understand in detail the advantages and disadvantages of using MAE loss.

\begin{figure}[h!]
    \centering
    \includegraphics[width=8cm]{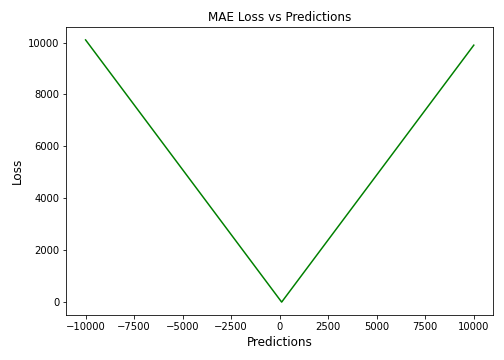}
    \caption{Performance Graph of MAE Loss vs Predictions}
    \label{fig:galaxy}
\end{figure}

\begin{table}[htbp]
\begin{small}
\caption{Advantages and Disadvantages of Using Mean Absolute Error Loss Function \cite{Qi_2020}.}
\begin{center}
\begin{tabular}{|p{3.75cm}|p{3.75cm}|}
\hline
\textbf{Advantages} & {\textbf{Disadvantages}} \\

\hline
MAE is computationally cheap because of its simplicity and provides an even measure of how well the model performs. & MAE follows a linear scoring approach, which means that all errors are weighted equally when computing the mean. Because of the steepness of MAE, we may hop beyond the minima during backpropagation.\\ \hline

MAE is less sensitive towards outliers &  MAE is not differentiable at zero, therefore it can be challenging to compute gradients.\\ \hline

\end{tabular}
\label{table1}
\end{center}
\end{small}
\end{table}


\subsection{Mean Squared Error (MSE)}

Squared Error Loss, also known as L2 loss, is the squared difference between the prediction and actual values.

$$ L2 = (y_{actual} - y_{predicted})^2 $$

The cost function for Squared Error is called the Mean of Squared Error. When it comes to regression loss functions, practically every data scientist prefers Mean Squared Error.

$$ MSE = \frac{1}{N} \sum_{i=1}^{N}(y_i - \hat{y_i})^2 $$

Here, $N$ is the number of data samples, $y_i$ is the true value, and $\hat{y_i}$ is the predicted value.

MSE is also referred to as a quadratic loss because the penalty is squared rather than directly proportional to the error. The outliers are given more weight when the error is squared, creating a smooth gradient for minor errors. Optimization algorithms benefit from this penalization for huge errors as it helps obtain the optimum values for parameters. Given that the mistakes are squared, MSE can never be negative, and the error's value can be anywhere between 0 and infinity. With increasing mistakes, MSE grows exponentially, and the MSE value of a good model will be close to zero \cite{allen1971mean}. Refer to Table \ref{table2} to learn in detail about the advantages and disadvantages of using MSE loss.

\begin{figure}[h!]
    \centering
    \includegraphics[width=8cm]{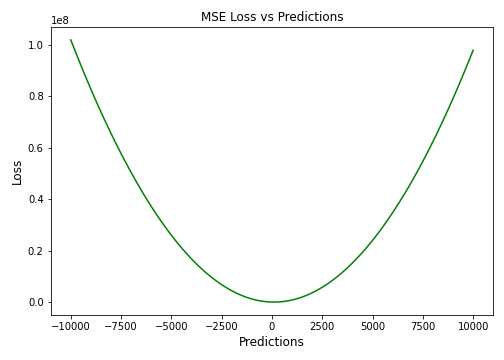}
    \caption{Performance Graph of MSE Loss vs Predictions}
    \label{fig:galaxy}
\end{figure}

\begin{table}[htbp]
\begin{small}
\caption{Advantages and Disadvantages of Using Mean Squared Error Loss Function \cite{allen1971mean}.}
\begin{center}
\begin{tabular}{|p{3.75cm}|p{3.75cm}|}
\hline
\textbf{Advantages} & {\textbf{Disadvantages}} \\

\hline

MSE aids in the efficient convergence to minima for tiny mistakes as the gradient gradually decreases. & Squaring the values accelerates the rate of training, but a higher loss value may result in a substantial leap during back propagation, which is undesirable.\\ \hline

MSE values are expressed in quadratic equations, aids to penalizing model in case of outliers. & MSE is especially sensitive to outliers, which means that significant outliers in data may influence our model performance. \\ \hline

\end{tabular}
\label{table2}
\end{center}
\end{small}
\end{table}


\subsection{Mean Bias Error (MBE)}

A measuring procedure propensity to overestimate or underestimate the value of a parameter is known as bias or Mean Bias Error. The only possible direction for bias is either a positive or negative one. A positive bias indicates that the data error is overstated, whereas a negative bias indicates that the error is underestimated. 

The discrepancy between the actual and anticipated values is measured as the mean bias error (MBE). The average bias in the prediction is captured and quantified by MBE. Except for not taking into account absolute values, it is virtually identical to MAE. MBE should be treated cautiously because positive and negative faults may cancel one another out \cite{kim2017mean}.

$$ MBE = \frac{1}{N} \sum_{i=1}^{N}(y_i - \hat{y_i}) $$

Here, $N$ is the number of data samples, $y_i$ is the true value, and $\hat{y_i}$ is the predicted value.

Refer to Table \ref{table3} to learn in detail about the advantages and disadvantages of using MBE loss.

\begin{figure}[h!]
    \centering
    \includegraphics[width=8cm]{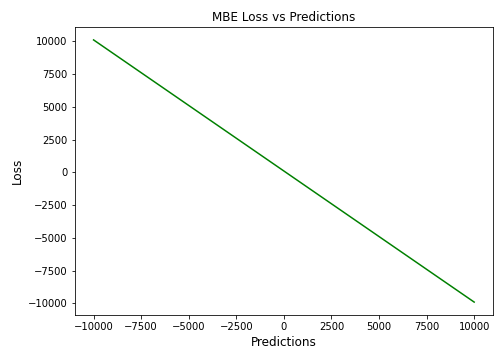}
    \caption{Performance Graph of MBE Loss vs Predictions}
    \label{fig:galaxy}
\end{figure}

\begin{table}[htbp]
\begin{small}
\caption{Advantages and Disadvantages of Using Mean Bias Error Loss Function \cite{kim2017mean}.}
\begin{center}
\begin{tabular}{|p{3.75cm}|p{3.75cm}|}
\hline
\textbf{Advantages} & {\textbf{Disadvantages}} \\

\hline

If you wish to identify and correct model bias, you should use MBE to determine the direction of the model (i.e., whether it is biased positively or negatively). & MBE tends to err in one direction continuously, while attempting to anticipate traffic patterns. Given that the errors tend to cancel each other out, it is not a suitable loss function for numbers ranging from $(-\infty, \infty)$ .  \\ \hline
\end{tabular}
\label{table3}
\end{center}
\end{small}
\end{table}


\subsection {Relative Absolute Error (RAE)}

The calculation of relative absolute error involves taking the total absolute error and dividing it by the absolute difference between the mean and the actual value.

RAE is expressed as,

$$ RAE = \frac{\sum_{i=1}^{N}|y_i - \hat{y_i}|}{\sum_{i=1}^{N}|y_i - \Bar{y}|} $$

where $$ \Bar{y} = \frac{1}{N} \sum_{i=1}^{N}(y_i) $$

Here, $N$ is the number of data samples, $y_i$ is the true value, $\hat{y_i}$ is the predicted value, and $\Bar{y}$ is the mean of $N$ actual values.

RAE is a ratio-based metric that assesses the efficacy of a predictive model. RAE has a possible value between 0 and 1. Values near zero, with zero being the best value, are characteristics of a good model. This error demonstrates how the target function's mean deviation from its mean and the mean residual relate to one another \cite{reich2016case}.

Refer to Table \ref{table4} to learn in detail about the advantages and disadvantages of using RAE loss.

\begin{figure}[h!]
    \centering
    \includegraphics[width=8cm]{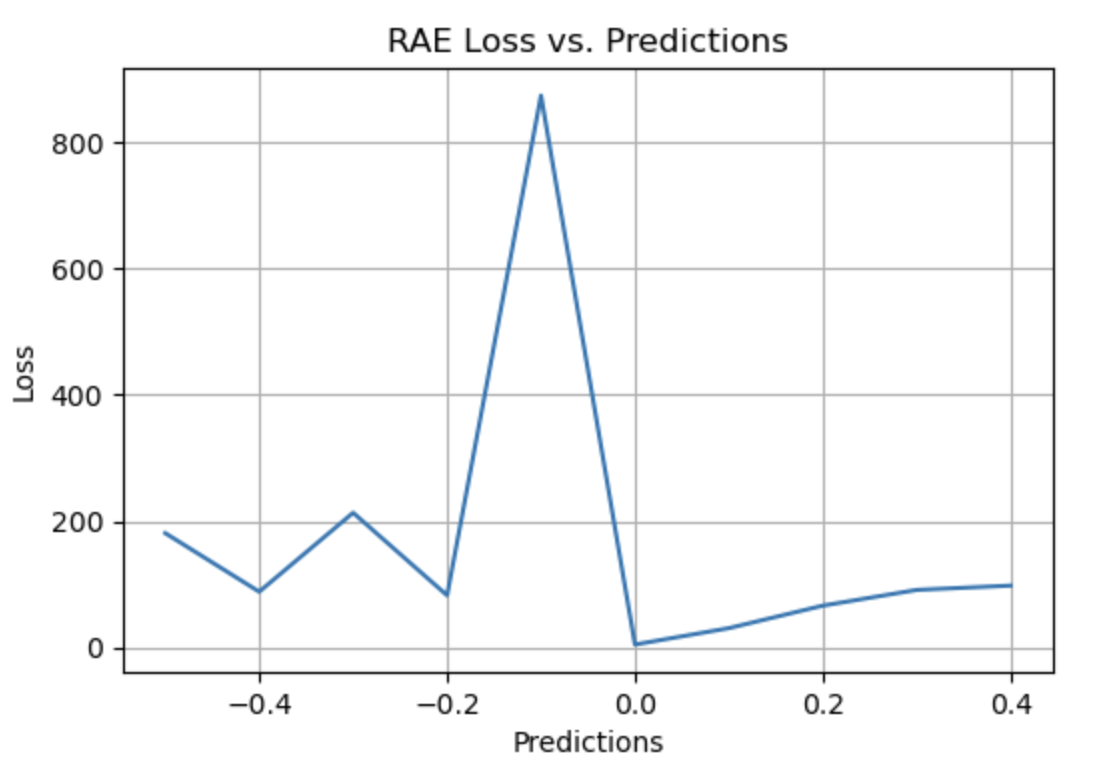}
    \caption{Performance Graph of RAE Loss vs Predictions}
    \label{fig:galaxy}
\end{figure}

\begin{table}[htbp]
\begin{small}
\caption{Advantages and Disadvantages of Using Relative Absolute Error Loss Function \cite{reich2016case}.}
\begin{center}
\begin{tabular}{|p{3.75cm}|p{3.75cm}|}
\hline
\textbf{Advantages} & {\textbf{Disadvantages}} \\

\hline

RAE can compare models where errors are measured in different units. & If the reference forecast is equal to the ground truth, RAE can become undefinable, which is one of its key drawbacks. \\ \hline

\end{tabular}
\label{table4}
\end{center}
\end{small}
\end{table}


\subsection{Relative Squared Error (RSE)}

The relative squared error (RSE) measures how inaccurate the results would have been without a simple predictor. This straightforward predictor merely represents the average of the actual values. As a result, the relative squared error divides the total squared error by the total squared error of the simple predictor to normalize it. It is possible to compare it among models whose errors are calculated in various units \cite{arnold2004some}.

The equation determines the relative squared error-

$$ RSE = \frac{\sum_{i=1}^{N}(y_i - \hat{y_i})^2}{\sum_{i=1}^{N}(y_i - \Bar{y})^2} $$

where $$ \Bar{y} = \frac{1}{N} \sum_{i=1}^{N} y_i $$

Here, $N$ is the number of data samples, $y_i$ is the true value, $\hat{y_i}$ is the predicted value, and $\Bar{y}$ is the mean of $N$ actual values.

Refer to Table \ref{table5} to learn in detail about the advantages and disadvantages of using RSE loss.

\begin{figure}[h!]
    \centering
    \includegraphics[width=8cm]{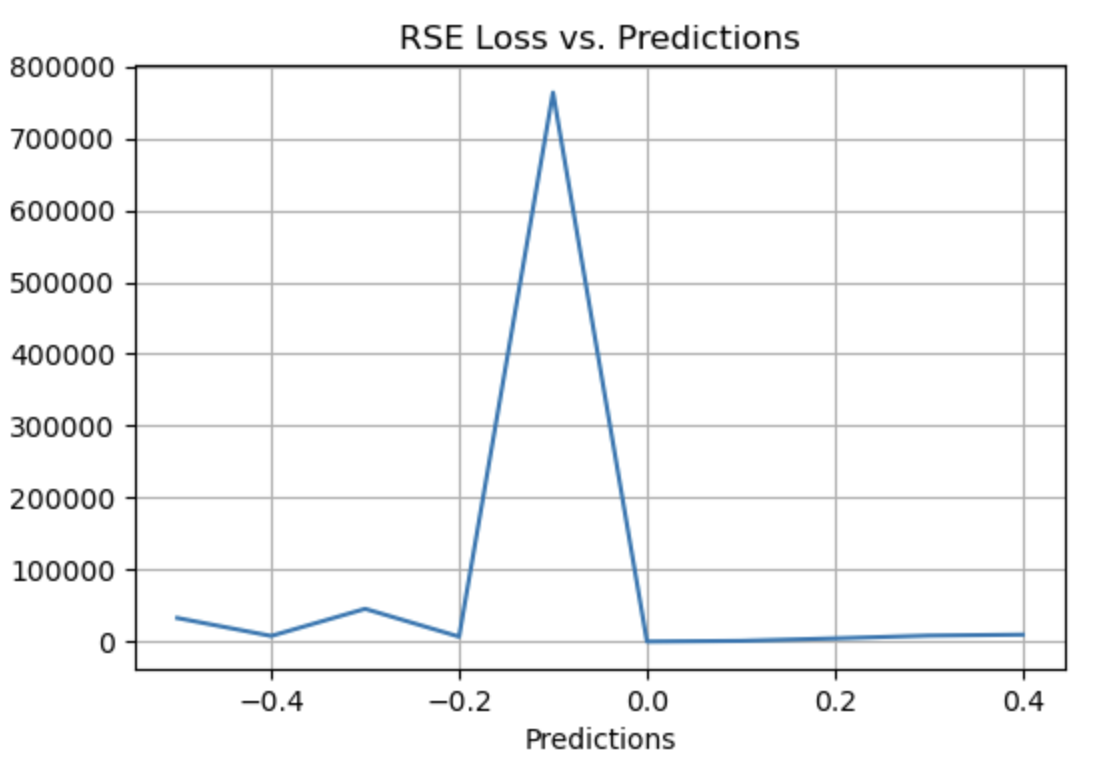}
    \caption{Performance Graph of RSE Loss vs Predictions}
    \label{fig:galaxy}
\end{figure}

\begin{table}[htbp]
\begin{small}
\caption{Advantages and Disadvantages of Using Relative Squared Error Loss Function \cite{arnold2004some}.}
\begin{center}
\begin{tabular}{|p{3.75cm}|p{3.75cm}|}
\hline
\textbf{Advantages} & {\textbf{Disadvantages}} \\

\hline
RSE is independent of scale. It can compare models when errors are measured in various units. & RSE is not affected by the predictions' mean or size.\\ \hline

\end{tabular}
\label{table5}
\end{center}
\end{small}
\end{table}


\subsection {Mean Absolute Percentage Error (MAPE)}

The mean absolute percentage error (MAPE), also known as the mean absolute percentage deviation (MAPD), is a metric used to assess the accuracy of a forecast system. It calculates the average absolute percent error as a percentage for each time period by subtracting actual values from actual values divided by actual values. Absolute errors, regardless of sign, are used to calculate percentage errors, and positive and negative errors are prevented from canceling one other out.

Because the variable's units are scaled to percent units, the mean absolute percentage error (MAPE) is widely used for forecasting errors. When there are no outliers in the data, it works well and is often used in regression analysis and model evaluation \cite{de2016mean}.

$$ MAPE = \frac{1}{N} \sum_{i=1}^{N}\frac{|y_i - \hat{y_i}|}{y_i}.100 \% $$
Here, $N$ is the number of data samples, $y_i$ is the true value, and $\hat{y_i}$ is the predicted value.

Refer to Table \ref{table6} to learn in detail about the advantages and disadvantages of using MAPE loss.

\begin{figure}[h!]
    \centering
    \includegraphics[width=8cm]{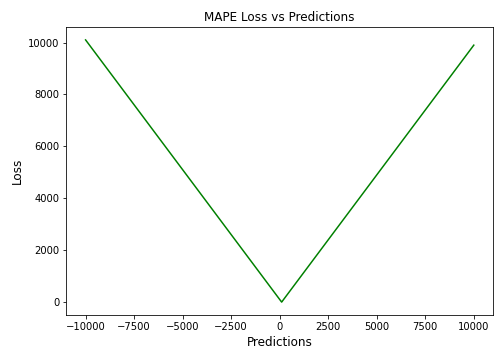}
    \caption{Performance Graph of MAPE Loss vs Predictions}
    \label{fig:galaxy}
\end{figure}

\begin{table}[htbp]
\begin{small}
\caption{Advantages and Disadvantages of Using Mean Absolute Percentage Error Loss Function \cite{de2016mean}.}
\begin{center}
\begin{tabular}{|p{3.75cm}|p{3.75cm}|}
\hline
\textbf{Advantages} & {\textbf{Disadvantages}} \\

\hline

MAPE Loss is computed by standardizing all errors on a single scale of hundred. & As denominator of the MAPE equation is the predicted output, which can be zero leading to undefined value. \\ \hline

As the error estimates are expressed in percentages, MAPE is independent of the scale of the variables. The issue of positive numbers canceling out negative ones is avoided since MAPE utilizes absolute percentage mistakes. & Positive errors are penalized less by MAPE than negative ones. Therefore, it is biased when we compare the precision of prediction algorithms since it defaults to selecting one whose results are too low.\\ \hline

\end{tabular}
\label{table6}
\end{center}
\end{small}
\end{table}


\subsection {Root Mean Squared Error (RMSE)}

The square root of MSE is used to calculate RMSE. The Root Mean Square Deviation is another name for RMSE. It accounts for variations from the actual value and measures the average magnitude of the errors. RMSE can be applied to a variety of features because it aids in determining whether a feature enhances model prediction or not.

$$ RMSE = \sqrt{\frac{1}{N} \sum_{i=1}^{N}(y_i - \hat{y_i})^2} $$

Here, $N$ is the number of data samples, $y_i$ is the true value, and $\hat{y_i}$ is the predicted value.

RMSE is most useful when huge errors are incredibly undesirable. According to competition regulations, submissions are judged based on the logarithm of the anticipated value and the logarithm of the observed sales price. Using logs means that errors in estimating expensive and inexpensive dwellings will impact the outcome.

The RMSE, which offers the average model prediction error in units of the variable of interest, indicates the model's absolute fit to the data. The model's performance and predictions improve with decreasing RMSE, and a greater RMSE denotes a significant departure from the ground truth in the residual \cite{chai2014root}.

Refer to Table \ref{table7} to learn in detail about the advantages and disadvantages of using RMSE loss.

\begin{figure}[h!]
    \centering
    \includegraphics[width=8cm]{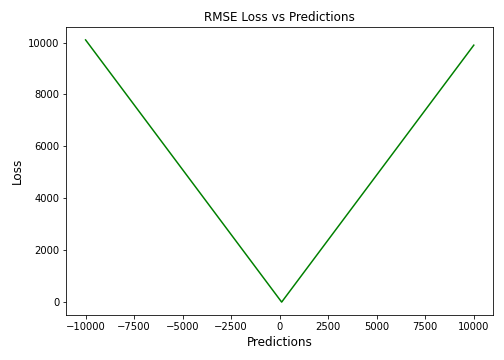}
    \caption{Performance Graph of RMSE Loss vs Predictions}
    \label{fig:galaxy}
\end{figure}

\begin{table}[htbp]
\begin{small}
\caption{Advantages and Disadvantages of Using Root Mean Squared Error Loss Function \cite{chai2014root}.}
\begin{center}
\begin{tabular}{|p{3.75cm}|p{3.75cm}|}
\hline
\textbf{Advantages} & {\textbf{Disadvantages}} \\
\hline

RMSE works as a training heuristic for models. Many optimization methods choose it because it is easily differentiable and computationally straightforward. & As RMSE is still a linear scoring function, the gradient is abrupt around minima. \\ \hline

Even with larger values, there are fewer extreme losses, and the square root causes RMSE to penalize errors less than MSE. & The scale of data determines the RMSE, as the errors' magnitude grows, so does sensitivity to outliers. In order to converge the model the sensitivity must be reduced, leading to extra overhead to use RMSE. \\ \hline

\end{tabular}
\label{table7}
\end{center}
\end{small}
\end{table}


\subsection{Mean Squared Logarithmic Error (MSLE)}

The mean squared logarithmic error (MSLE) measures the difference between actual and expected values.

Adding the logarithm has reduced MSLE concern for the percentual difference between the actual and forecasted values and the relative difference between the two.
Accordingly, MSLE will roughly treat tiny discrepancies between small actual and anticipated values and huge disparities between large true and forecasted values \cite{massmann2012analysing}.

The loss is the mean over the observed data of the squared differences between the true and predicted values after log transformation, or you can write it as a formula.

$$ MSLE = \frac{1}{N} \sum_{i=0}^{N}(log(y_i+1) - log(\hat{y_i}+1))^2 $$

Here, $N$ is the number of data samples, $y_i$ is the true value, and $\hat{y_i}$ is the predicted value. 

This loss can be interpreted as a measure of the ratio between the true and predicted values since:

$$ log(y_i+1) - log(\hat{y_i}+1) = log(\frac{y_i+1}{\hat{y_i}+1}) $$

Log(0) is not defined, but both y and can be defined mathematically. Therefore one is appended to both y and for mathematical convenience.

Refer to Table \ref{table8} to learn in detail about the advantages and disadvantages of using MSLE loss.

\begin{figure}[h!]
    \centering
    \includegraphics[width=8cm]{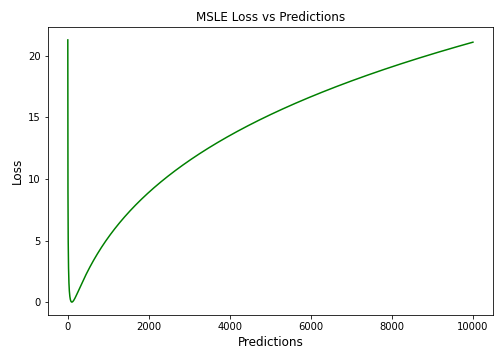}
    \caption{Performance Graph of MSLE Loss vs Predictions}
    \label{fig:galaxy}
\end{figure}

\begin{table}[htbp]
\begin{small}
\caption{Advantages and Disadvantages of Using Mean Squared Logarithmic Error Loss Function \cite{massmann2012analysing}.}
\begin{center}
\begin{tabular}{|p{3.75cm}|p{3.75cm}|}
\hline
\textbf{Advantages} & {\textbf{Disadvantages}} \\

\hline
Treats small differences between small actual and predicted values the same as big differences between large actual and predicted values. & MSLE penalizes underestimates more than overestimates.\\ \hline

\end{tabular}
\label{table8}
\end{center}
\end{small}
\end{table}


\subsection {Root Mean Squared Logarithmic Error (RMSLE)}

The root mean squared logarithmic error is determined by applying log to the actual and predicted numbers and then subtracting them. RMSLE is resistant to outliers when both minor and large errors are considered.

$$ RMSLE = \sqrt{\frac{1}{N} \sum_{i=0}^{N}(log(y_i+1) - log(\hat{y_i}+1))^2} $$

Here, $N$ is the number of data samples, $y_i$ is the true value, and $\hat{y_i}$ is the predicted value.

If the predicted value is less than the actual value, it penalizes the model more; if it is greater than the actual value, it penalizes the model less. High errors brought on by the log are not punished. As a result, the model has a far higher penalty for underestimating than for overestimating. This is useful in circumstances where we do not mind overestimation but underestimating is unacceptable \cite{mir2022improved}.

Refer to Table \ref{table9} to learn in detail about the advantages and disadvantages of using RMSLE loss.

\begin{figure}[h!]
    \centering
    \includegraphics[width=8cm]{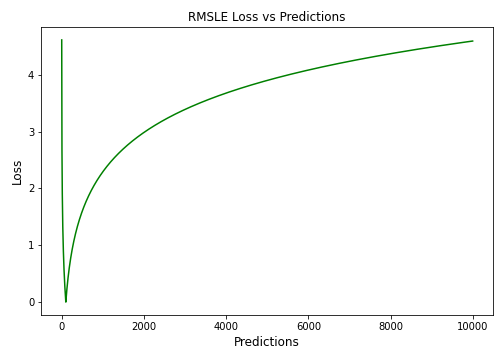}
    \caption{Performance Graph of RMSLE Loss vs Predictions}
    \label{fig:galaxy}
\end{figure}

\begin{table}[htbp]
\begin{small}
\caption{Advantages and Disadvantages of Using Root Mean Squared Logarithmic Error Loss Function \cite{mir2022improved}.}
\begin{center}
\begin{tabular}{|p{3.75cm}|p{3.75cm}|}
\hline
\textbf{Advantages} & {\textbf{Disadvantages}} \\

\hline

RMSLE is applicable at several scales and is not scale-dependent. It is unaffected by significant outliers. Only the relative error between the actual value and the anticipated value is taken into account. & Due to RMSLE's biased penalty, underestimating is penalized more severely than overestimation. \\ \hline

\end{tabular}
\label{table9}
\end{center}
\end{small}
\end{table}


\subsection{Normalized Root Mean Squared Error (NRMSE)}

The Normalized Root Mean Square Error (NRMSE) the RMSE facilitates the comparison between models with different scales. The variable has the observed range's normalized RMSE (NRMSE), which connects the RMSE to the observed range. As a result, it is possible to interpret the NRMSE as a portion of the whole range that the model resolves typically \cite{shcherbakov2013survey}.

$$ NRMSE = \frac{\sqrt{\frac{1}{N} \sum_{i=1}^{N}(y_i - \hat{y_i})^2}}{\Bar{o}} $$

Here, $N$ is the number of data samples, $y_i$ is the true value and, $\hat{y_i}$ is the predicted value, $\Bar{o}$ is the average of the observation value.

Refer to Table \ref{table10} to learn in detail about the advantages and disadvantages of using NRMSE loss.

\begin{figure}[h!]
    \centering
    \includegraphics[width=8cm]{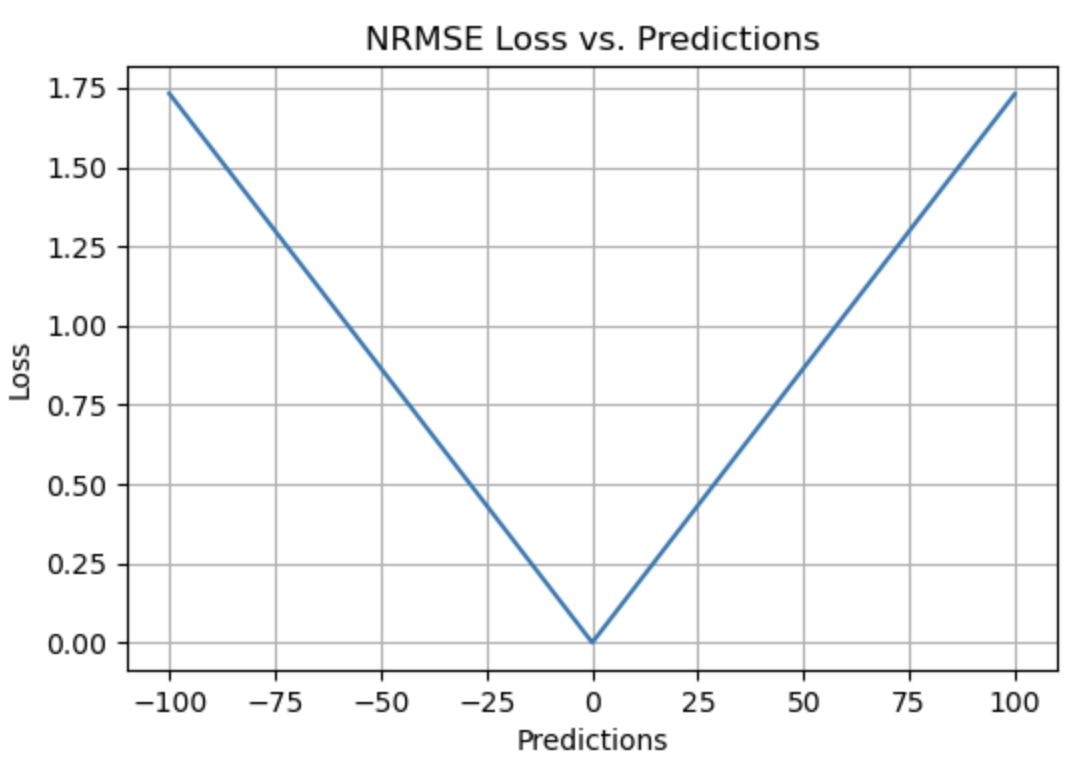}
    \caption{Performance Graph of NRMSE Loss vs Predictions}
    \label{fig:galaxy}
\end{figure}

\begin{table}[htbp]
\begin{small}
\caption{Explaining Advantages and Disadvantages of Using Normalized Root Mean Squared Error Loss Function \cite{shcherbakov2013survey}.}
\begin{center}
\begin{tabular}{|p{3.75cm}|p{3.75cm}|}
\hline
\textbf{Advantages} & {\textbf{Disadvantages}} \\

\hline
NRMSE overcomes the scale dependency and eases comparison between models of different scales or datasets. & NRMSE loses the units associated with the response variable. \\ \hline

\end{tabular}
\label{table10}
\end{center}
\end{small}
\end{table}


\subsection{Relative Root Mean Squared Error (RRMSE)}

RRMSE is an RMSE variant without dimensions. Relative Root Mean Square Error (RRMSE) is a root mean square error measure that has been scaled against the actual value and then normalized by the root mean square value. While the original measurements' scale constraints RMSE, RRMSE can be used to compare various measurement approaches. An enhanced RRMSE happens when your predictions turn out to be wrong, and the error is expressed by RRMSE either relatively or as a percentage \cite{chambers1986estimating}.

RRMSE Model accuracy is excellent when the model score is below 10\%, good when the model score is between  10\% to 20\%, fair when the model score is between 20\% to 30\%, and poor when the model score is greater than 30\%

RRMSE can be expressed as 

$$ RRMSE = \sqrt{\frac{\frac{1}{N}\sum_{i=1}^{N}(y_i - \hat{y_i})^2}{\sum_{i=1}^{N}(\hat{y_i})^2}} $$

Here, $N$ is the number of data samples, $y_i$ is the true value, and $\hat{y_i}$ is the predicted value. 

Refer to Table \ref{table11} to learn in detail about the advantages and disadvantages of using RRMSE loss.

\begin{figure}[h!]
    \centering
    \includegraphics[width=8cm]{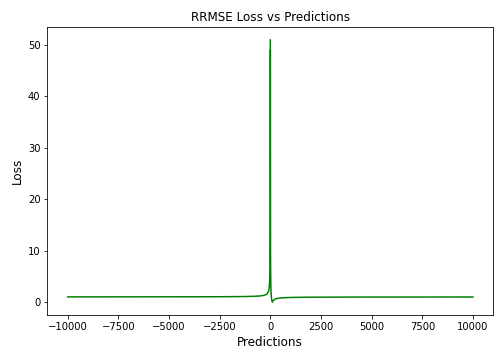}
    \caption{Performance Graph of RRMSE Loss vs Predictions}
    \label{fig:galaxy}
\end{figure}

\begin{table}[htbp]
\begin{small}
\caption{Explaining Advantages and Disadvantages of Using Relative Root Mean Squared Error Loss Function \cite{chambers1986estimating}.}
\begin{center}
\begin{tabular}{|p{3.75cm}|p{3.75cm}|}
\hline
\textbf{Advantages} & {\textbf{Disadvantages}} \\

\hline

RRMSE can be used to compare different measurement techniques. & RRMSE can hide inaccuracy in experiment results\\ \hline

\end{tabular}
\label{table11}
\end{center}
\end{small}
\end{table}


\subsection {Huber Loss}

Huber loss is an ideal combination of quadratic and linear scoring algorithms. There is also the hyperparameter delta ($\delta$). Loss is linear for values more than delta; for values less than delta, the loss is quadratic. This parameter distinguishes the Huber loss and is changeable based on the data.

$$ L_{\delta}= \left\{\begin{matrix} \frac{1}{2}(y - \hat{y})^{2} & if \left | (y - \hat{y})  \right | < \delta\\ \delta ((y - \hat{y}) - \frac1 2 \delta) & otherwise \end{matrix}\right.
$$

Here, $N$ is the number of data samples, $y_i$ is the true value, $\hat{y_i}$ is the predicted value, and ($\delta$) is the hyperparameter delta.

This equation effectively states that for loss values less than the delta, the MSE should be used; for loss values greater than the delta, the MAE should be used. This successfully combines the greatest features of both loss functions. 

Using the MAE for more significant loss numbers reduces the weight placed on outliers while still producing a well-rounded model. Simultaneously, the MSE is employed for the lesser loss values to keep a quadratic function around the center \cite{meyer2021alternative}.

Refer to Table \ref{table12} to learn in detail about the advantages and disadvantages of using Huber loss.

\begin{figure}[h!]
    \centering
    \includegraphics[width=8cm]{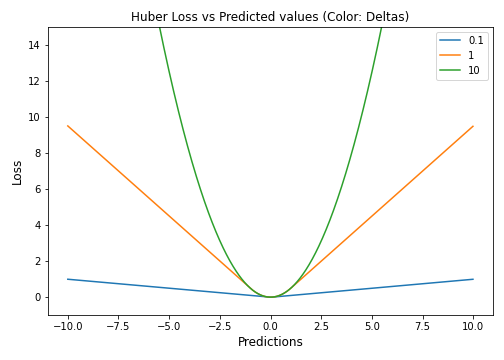}
    \caption{Performance Graph of Huber Loss vs Predictions}
    \label{fig:galaxy}
\end{figure}

\begin{table}[htbp]
\begin{small}
\caption{Explaining Advantages and Disadvantages of Using Huber Loss Function \cite{meyer2021alternative}.}
\begin{center}
\begin{tabular}{|p{3.75cm}|p{3.75cm}|}
\hline
\textbf{Advantages} & {\textbf{Disadvantages}} \\

\hline
Linearity above the delta guarantees that outliers are given appropriate weightage (Not as extreme as in MSE). Addition of hyper parameter delta ($\delta$) allows flexibility to adapt to any distribution & Huber loss is computationally expensive due to the additional conditionals and comparisons, especially if your dataset is huge. \\ \hline

The curved form below the delta guarantees that the steps are the correct length during backpropagation. & To achieve the best outcomes must be optimized, which raises training requirements. \\ \hline

\end{tabular}
\label{table12}
\end{center}
\end{small}
\end{table}


\subsection {LogCosh Loss }

Log cosh calculates the logarithm of the hyperbolic cosine of the error. More smoothly than quadratic loss, this function. It functions similarly to MSE but is unaffected by significant prediction errors. Given that it uses linear and quadratic scoring techniques, it is extremely close to Huber loss.

$$ Logcosh(t) = \sum_{i=1}^{N} log(cosh(\hat{y_i} - y_i)) $$

Here, $N$ is the number of data samples, $y_i$ is the true value, and $\hat{y_i}$ is the predicted value.

It has the distinction of being dual differentiable. Some optimization approaches, such as XGBoost, favor such functions over Huber functions, which can only be differentiated once. Log-Cosh computes the log of the hyperbolic cosine of the mistake \cite{saleh2022statistical}.

Refer to Table \ref{table13} to learn in detail about the advantages and disadvantages of using LogCosh loss.

\begin{figure}[h!]
    \centering
    \includegraphics[width=8cm]{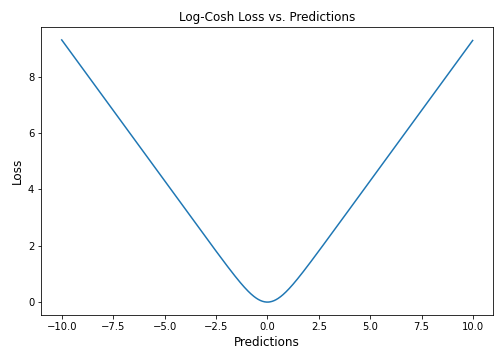}
    \caption{Performance Graph of LogCosh Loss vs Predictions}
    \label{fig:galaxy}
\end{figure}

\begin{table}[htbp]
\begin{small}
\caption{Explaining Advantages and Disadvantages of Using Log-Cosh Loss Function \cite{saleh2022statistical}.}
\begin{center}
\begin{tabular}{|p{3.75cm}|p{3.75cm}|}
\hline
\textbf{Advantages} & {\textbf{Disadvantages}} \\

\hline

As Log-cosh calculates the log of hyperbolic cosine of the error. Therefore, it has a considerable advantage over Huber loss for its’ property of continuity and differentiability. & It is less adaptable than Huber since it operates on a fixed scale (no $\delta$). \\ \hline

There are fewer computations required compared with Huber. & The derivation is more challenging than Huber loss and necessitates more research. \\ \hline

\end{tabular}
\label{table13}
\end{center}
\end{small}
\end{table}

\begin{table*}[!h]
\begin{center}
\caption{Summary of Loss Functions}

\begin{tabular} {|l|p{10cm}|}
\hline

\textbf{Loss Function} & {\textbf{Use Cases}} \\
\hline

Mean Absolute Error (MAE) & If your use case demands an error metric that treats all errors equally and returns a more interpretable value.  \\ \hline

Mean Squared Error (MSE) & If the outliers represent significant anomalies, your use case demands an error metric where outliers should be detected.\\ \hline

Mean Bias Error (MBE) & The directionality of bias can be preserved with MBE. If you are simply concerned with a system's net or cumulative behavior, MBE can help you assess how well biases balance out. \\ \hline

Relative Absolute Error (RAE) & RAE is a method to measure the performance of a predictive model. If you are simply concerned and want a metric that indicates and compares how well a model works. \\ \hline

Relative Squared Error (RSE) & RSE is not scale-dependent. If your use case demands similarity between models, it can be used to compare models where errors are measured in different units. \\ \hline

Mean Absolute Percentage Error (MAPE) & Since its error estimates are expressed in percentages. It is independent of the scale of the variables. MAPE can be used if your use case requires that all mistakes be normalized on a standard scale. \\ \hline

Root Mean Squared Error (RMSE) & We can utilize RMSE if your use case necessitates a computationally straightforward and easily differentiable loss( as many optimization techniques do). Also, it does not penalize errors as severely as MSE does.\\ \hline

Mean Squared Logarithmic Error (MSLE) & MSLE reduce the punishing effect of significant differences in large predicted values. When the model predicts unscaled quantities directly, it may be more appropriate as a loss measure. \\ \hline

Root Mean Squared Logarithmic Error (RMSLE) & RMSLE has a significant penalty for underestimation than overestimation. If your use case requires situations where we are not bothered by overestimation, but underestimation is not acceptable, we can use RMSLE. \\ \hline

Normalized Root Mean Squared Error (NRMSE) & When comparing models with different dependent variables or when the dependent variables are changed, NRMSE is a valuable measure (log-transformed or standardized). It eliminates scale dependence and allows for easier comparison of models of different scales or even datasets.\\ \hline

Relative Root Mean Squared Error (RRMSE) & While the scale of the original measurements limits RMSE, RRMSE can be used if your use case necessitates comparing different measurement techniques. Also, RRMSE expresses the error relatively or in a percentage form.\\ \hline

Huber Loss & Huber Loss curves around the minima, which decreases the gradient and is more robust to outliers while optimally penalizing the incorrect values. \\ \hline

LogCosh Loss & Logcosh works similarly to mean squared error but is less affected by the occasional wildly incorrect prediction. It has all of the benefits of Huber loss; unlike Huber loss, It is twice differentiable everywhere. \\ \hline

Quantile Loss & Use Quantile Loss when predicting an interval instead of point estimates. Quantile Loss can also be used to calculate prediction intervals in neural nets and tree-based models, and It is robust to outliers. \\ \hline

\end{tabular}
\label{tab0}
\end{center}
\end{table*}


\subsection {Quantile Loss }

The Quantile regression loss function is used to forecast quantiles. The quantile is the value that indicates how many values in the group fall below or above a particular threshold. It calculates the conditional median or quantile of the response (dependent) variables across values of the predictor (independent) variables. Except for the 50th percentile, where it is MAE, the loss function is an extension of MAE. 

It makes no assumptions about the response's parametric distribution and offers prediction intervals even for residuals with non-constant variance.

$$ Quantile Loss = \sum_{i=y_i < \hat{y_i}}(\gamma -1)| y_i - \hat{y_i})| + \sum_{i=y_i \geq \hat{y_i}} (\gamma)| y_i - \hat{y_i})| $$

Here, $\gamma$ represents the required quantile, $y_i$ is the true value, and $\hat{y_i}$ is the predicted value. The quantile values are chosen based on how we wish to balance the positive and the negative errors.

The value of the loss function above ranges from 0 to 1. The first portion of the formula will predominate in cases of underestimation, whereas the second component will predominate in cases of overestimation. Different penalties for over- and under-prediction are given depending on the selected quantile value. 

Underestimation and overestimation are penalized by the same factor when = 0.5, and the median is then determined. Overestimation is punished more severely than underestimating when the value is greater. For instance, the model penalizes overestimation when = 0.75, costing three times as much as underestimating. Gradient descent-based optimization techniques learn from the quantiles rather than the mean \cite{koenker2017handbook}.

Refer to Table \ref{table14} to learn in detail about the advantages and disadvantages of using Quantile loss.

\begin{figure}[h!]
    \centering
    \includegraphics[width=8cm]{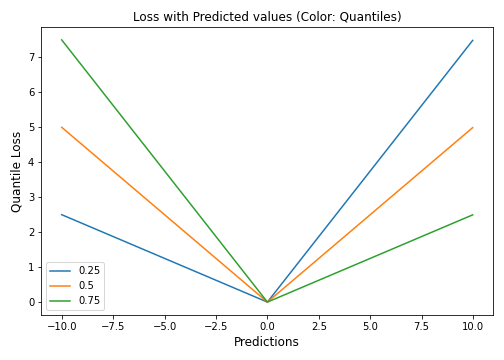}
    \caption{Performance Graph of Quantile Loss vs Predictions}
    \label{fig:galaxy}
\end{figure}

\begin{table}[htbp]
\begin{small}
\caption{Explaining Advantages and Disadvantages of Using Quantile Loss Function \cite{koenker2017handbook}.}
\begin{center}
\begin{tabular}{|p{3.75cm}|p{3.75cm}|}
\hline
\textbf{Advantages} & {\textbf{Disadvantages}} \\

\hline

It is robust to outliers. & Quantile loss is computationally intensive. \\ \hline

It is beneficial for making interval predictions as opposed to point estimates. This function can also be used in neural networks and tree-based models to determine prediction intervals. & Quantile loss will be worse if we estimate the mean or use a squared loss to quantify efficiency. \\ \hline

\end{tabular}
\label{table14}
\end{center}
\end{small}
\end{table}

\begin{table}[h!]
\begin{small}
\caption{Summary of Loss Functions Performance on Electricity Dataset}
\begin{center}
\begin{tabular}{|p{4cm}|c|c|c|}
\hline
\textbf{Loss}&\multicolumn{3}{|c|}{\textbf{Electricity Dataset}} \\
\cline{2-4} 
\textbf{Functions} & \textbf{\textit{P10}}& \textbf{\textit{P50}}& \textbf{\textit{P90}} \\
\hline
Mean Absolute Error (MAE) & 0.097  & 0.218  & 0.336 \\ \hline
\textbf{Mean Squared Error (MSE)} &\textbf{0.082}  & \textbf{0.149}  & \textbf{0.211} \\ \hline
Mean Bias Error (MBE) & 1068.7  & 579.25  & 119.86 \\ \hline
Relative Absolute Error (RAE) & 0.143  & 0.198  & 0.255 \\ \hline
Relative Squared Error (RSE) & 0.428  & 0.340  & 1.103 \\ \hline
Mean Absolute Percentage Error (MAPE) & 0.241  & 0.415  & 0.589 \\ \hline
Root Mean Squared Error (RMSE) & 0.135  & 0.234  & 0.325 \\ \hline
Mean Squared Logarithmic Error (MSLE) & 0.192  & 0.868  & 1.249 \\ \hline
Root Mean Squared Logarithmic Error (RMSLE) & 0.239  & 1.535  & 2.596 \\ \hline
Normalized Root Mean Squared Error (NRMSE) & 0.120  & 0.424  & 0.451 \\ \hline
\textbf{Relative Root Mean Squared Error (RRMSE)} & \textbf{0.188}  & \textbf{0.192}  & \textbf{0.194} \\ \hline
Huber Loss (delta = 0.5) &  0.126 & 0.203  & 0.268  \\ \hline
Log Cosh Loss & 0.125  & 0.227  & 0.318 \\ \hline
\textbf{Quantile Loss} & \textbf{0.073}  & \textbf{0.173}  & \textbf{0.074} \\ \hline
\end{tabular}
\label{table16}
\end{center}
\end{small}
\end{table}


\section{Experiments}

\subsection{Datasets}

\begin{itemize}
\item \textbf{Electricity Load Dataset} - Dataset contains electricity consumption of 370 points/clients.

\item \textbf{PEMS-SF Dataset} - Dataset contains 15 months worth of daily data (440 daily records) that describes the occupancy rate, between 0 and 1, of different car lanes of the San Francisco bay area freeways across time.

\item \textbf{Corporación Favorita Dataset} - Grocery dataset containing dates, store and item information, whether that item was being promoted, and the unit sales.

\item \textbf{Volatility Dataset} - Dataset contains daily non-parametric measures of how volatility financial assets or indexes were in the past.

\end{itemize}

Decomposition is mainly used for time series analysis, and as a tool for analysis, it can be utilized to guide forecasting models for your issue.

\begin{figure}[h!]
    \centering
    \includegraphics[width=8cm]{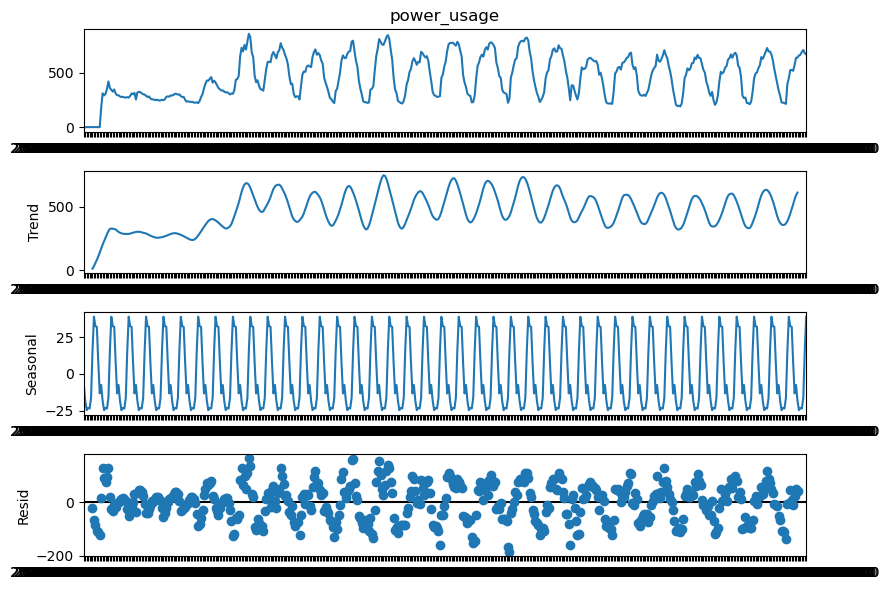}
    \caption{Sample Temporal Representation of Electricity Load Dataset.}
    \label{fig:galaxy}
\end{figure}

\begin{figure}[h!]
    \centering
    \includegraphics[width=8cm]{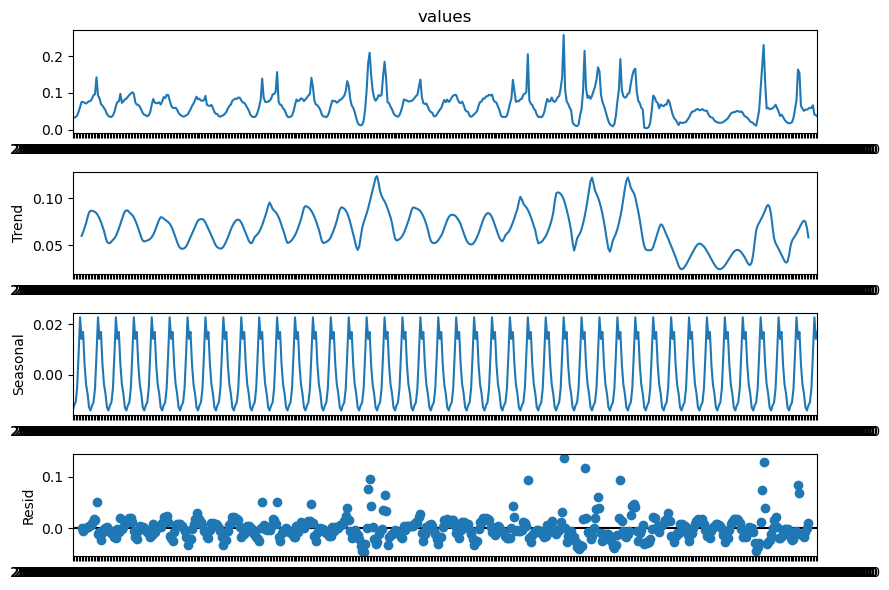}
    \caption{Sample Temporal Representation of PEMS-SF Dataset.}
    \label{fig:galaxy}
\end{figure}

\begin{figure}[h!]
    \centering
    \includegraphics[width=8cm]{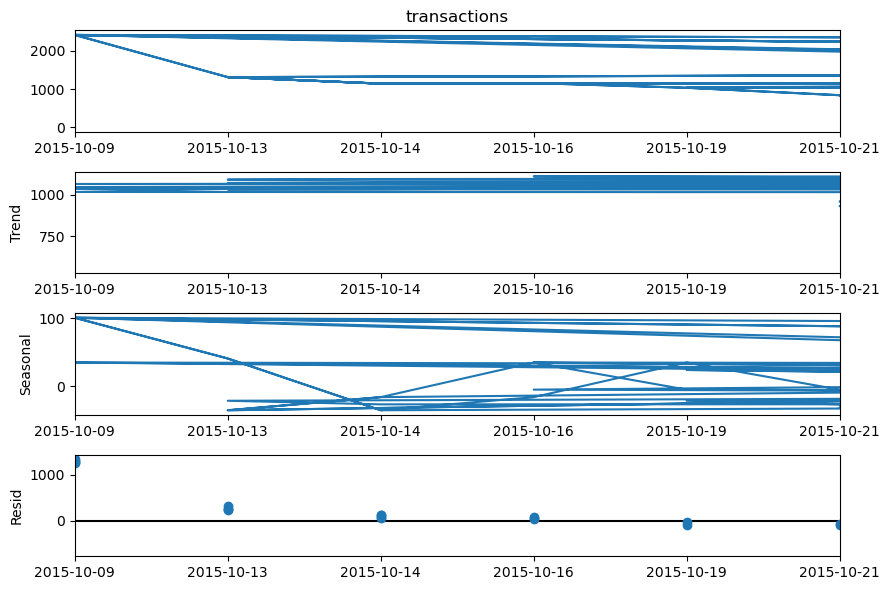}
    \caption{Sample Temporal Representation of Corporacion Favorita Dataset.}
    \label{fig:galaxy}
\end{figure}

\begin{figure}[h!]
    \centering
    \includegraphics[width=8cm]{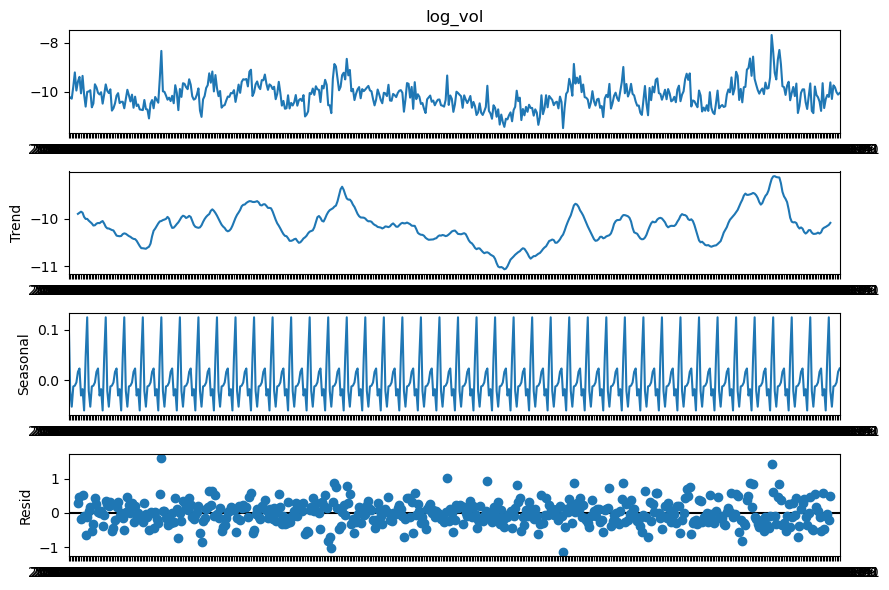}
    \caption{Sample Temporal Representation of Volatility Dataset.}
    \label{fig:galaxy}
\end{figure}

\subsection{Performance Metrics}
For this experiment setup, we evaluated the performance of all loss functions using P10, P50, and P90 metrics. 

An upper and lower bound for forecasts can be provided via quantiles. An 80\% confidence interval is a range of values that can be obtained, for instance, by utilizing the forecast types 0.1 (P10) and 0.9 (P90). In 10\% of cases, the observed value should be less than the P10 value, and in 90\% of cases, the P90 value should be higher. You can anticipate that the actual value will fall inside those ranges 80\% of the time by creating forecasts at p10 and P90. The shaded area between P10 and P90 in the following figure represents this range of values.

\begin{figure}[h!]
    \centering
    \includegraphics[width=8cm]{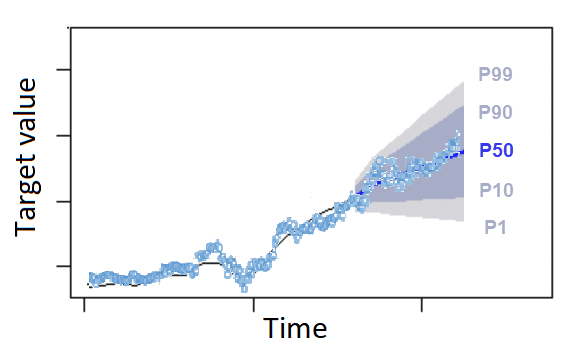}
    \caption{Illustration of Metrics used for evaluation of model performance.}
    \label{fig:galaxy}
\end{figure}

\begin{itemize}
\item \textbf{P10 (0.1)} - The true value is expected to be lower than the predicted value 10\% of the time.
\item \textbf{P50 (0.5)} - The true value is expected to be lower than the predicted value 50\% of the time. This is also known as the median forecast.
\item \textbf{P90 (0.9)} - The true value is expected to be lower than the predicted value 90\% of the time.
\end{itemize}

We have implemented Temporal Fusion Transformers \cite{fusion_transformers} architecture and performed experiments using parameters in \ref{table15}.

\begin{table}[h!]
\begin{small}
\caption{Experiment Parameters}
\begin{center}
\begin{tabular}{|p{3.75cm}|p{3.75cm}|}
\hline

\textbf{Parameter} & {\textbf{Value}} \\ 
\hline

dropout rate &  0.3  \\ \hline
hidden layer size &  16  \\ \hline
learning rate &  0.03  \\ \hline
max gradient norm &  0.01  \\ \hline
minibatch size &  64  \\ \hline
num heads &  1  \\ \hline
stack size &  1  \\ \hline
total time steps &  257  \\ \hline
num encoder steps &  252  \\ \hline
num epochs & 15  \\ \hline
early stopping patience & 5  \\ \hline

\end{tabular}
\label{table15}
\end{center}
\end{small}
\end{table}

We have used 1000 training and 100 validation samples and performed experiments using 14 loss functions. Other loss functions were either resolved into our chosen loss function or were not fit for the listed datasets. \\

Fig. 20-23 showcase the performance of respective loss functions on the above-mentioned datasets.

\begin{figure}[h!]
\centerline{\includegraphics[scale = 0.27]{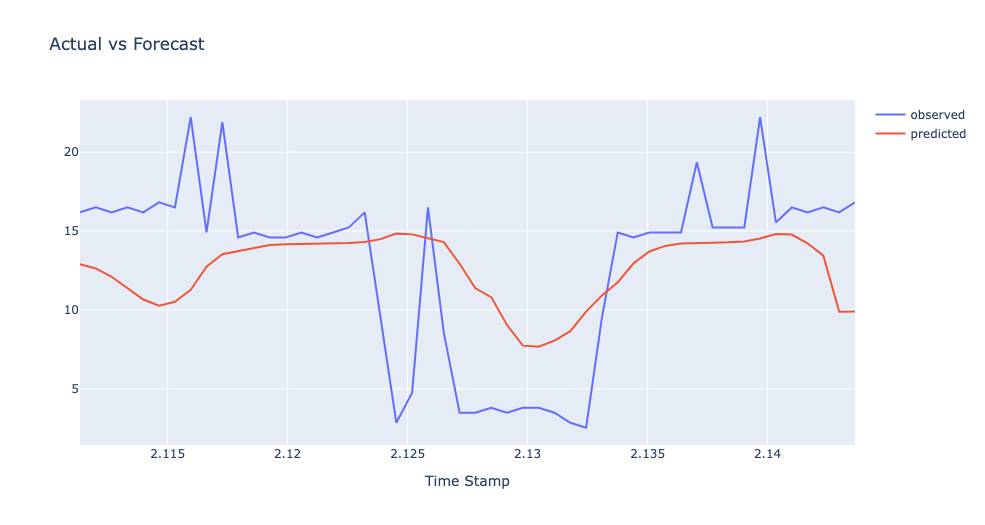}}
\caption{Observed(blue) vs Predicted(red) plot using Quantile Loss on Sample Electricity Dataset}
\label{fig}
\end{figure}

%
%



\begin{figure}[h!]
\centerline{\includegraphics[scale = 0.27]{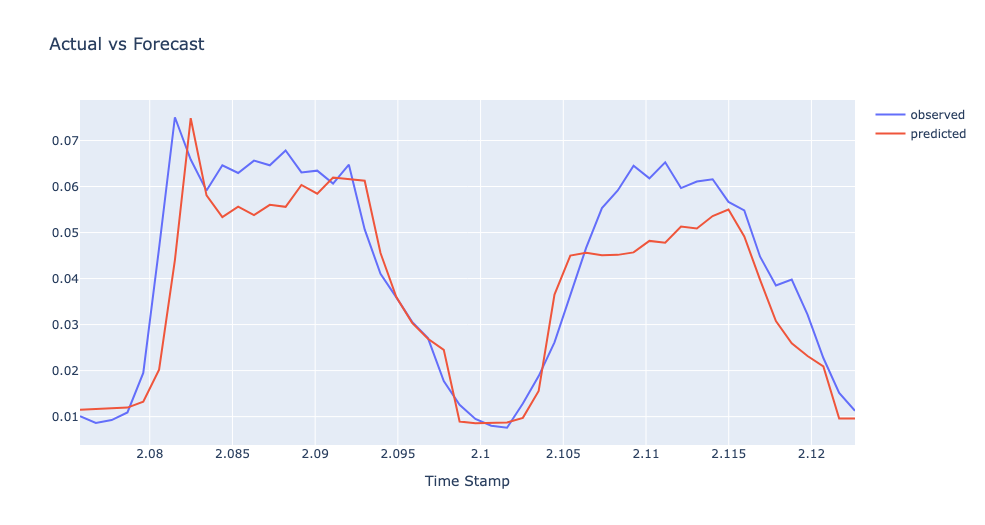}}
\caption{Observed(blue) vs Predicted(red) plot using Quantile Loss on Sample Traffic Dataset}
\label{fig}
\end{figure}




\begin{figure}[h!]
\centerline{\includegraphics[scale = 0.27]{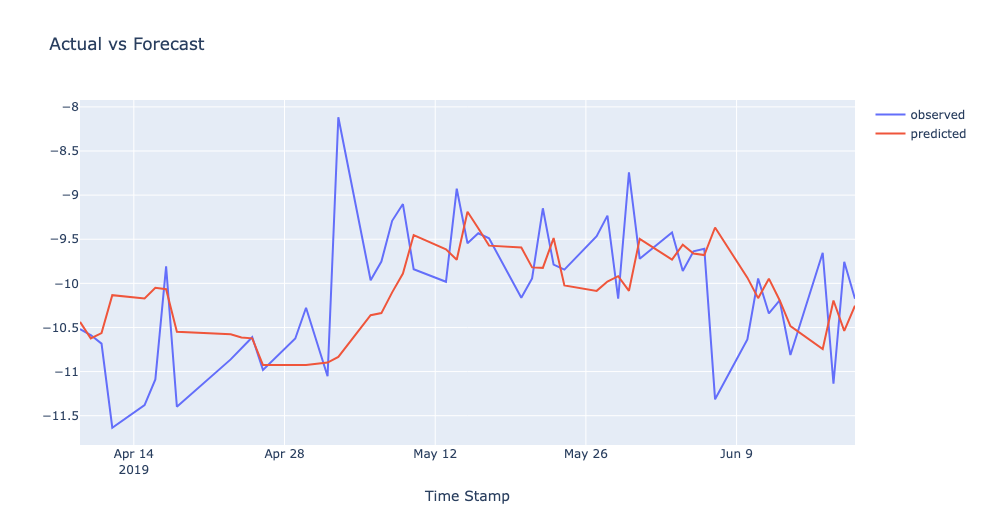}}
\caption{Observed(blue) vs Predicted(red) plot using Quantile Loss on Sample Favorita Dataset}
\label{fig}
\end{figure}



\begin{figure}[h!]
\centerline{\includegraphics[scale = 0.27]{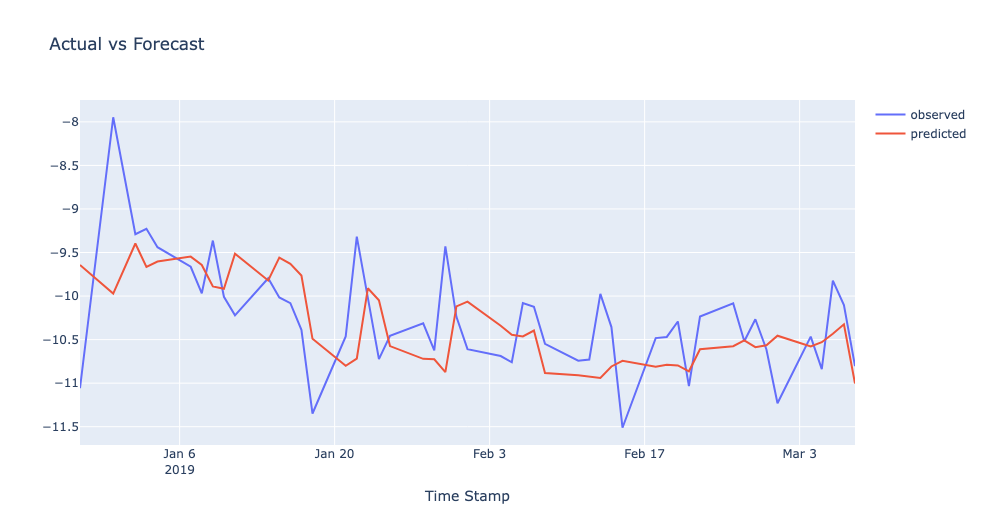}}
\caption{Observed(blue) vs Predicted(red) plot using Quantile Loss on Sample Volatility Dataset}
\label{fig}
\end{figure}

\begin{table}[h!]
\begin{small}
\caption{Summary of Loss Functions Performance on Traffic Dataset}
\begin{center}
\begin{tabular}{|p{4cm}|c|c|c|}
\hline
\textbf{Loss}&\multicolumn{3}{|c|}{\textbf{Traffic Dataset}} \\
\cline{2-4} 
\textbf{Functions} & \textbf{\textit{P10}}& \textbf{\textit{P50}}& \textbf{\textit{P90}} \\
\hline
\textbf{Mean Absolute Error (MAE)} & \textbf{0.214}  & \textbf{0.259}  & \textbf{0.302} \\ \hline
Mean Squared Error (MSE) & 0.339  & 0.312  & 0.334 \\ \hline
Mean Bias Error (MBE) & 1374.8  & 792.30  & 159.87 \\ \hline
Relative Absolute Error (RAE) & 0.279  & 0.328  & 0.394 \\ \hline
Relative Squared Error (RSE) & 0.812  & 0.619  & 0.510 \\ \hline
Mean Absolute Percentage Error (MAPE) & 0.315 & 0.431  & 0.516 \\ \hline
Root Mean Squared Error (RMSE) & 0.353  & 0.339  & 0.337 \\ \hline
Mean Squared Logarithmic Error (MSLE) & 0.570  & 0.485  & 4.200 \\ \hline
Root Mean Squared Logarithmic Error (RMSLE) & 0.507 & 2.077  & 3.581 \\ \hline
Normalized Root Mean Squared Error (NRMSE) & 0.662  & 0.685  & 0.502 \\ \hline
Relative Root Mean Squared Error (RRMSE) & 0.862  & 1.193  & 1.304 \\ \hline
Huber Loss (delta = 0.5) & 0.287  & 0.354  & 0.433 \\ \hline
 \textbf{Log Cosh Loss} & \textbf{0.212} & \textbf{0.248}  & \textbf{0.289} \\ \hline
 \textbf{Quantile Loss} & \textbf{0.108}  & \textbf{0.315}  & \textbf{0.232} \\ \hline
\end{tabular}
\label{tab1}
\end{center}
\end{small}
\end{table}

\begin{table}[h!]
\begin{small}
\caption{Summary of Loss Functions Performance on Favorita Dataset}
\begin{center}
\begin{tabular}{|p{4cm}|c|c|c|}
\hline
\textbf{Loss}&\multicolumn{3}{|c|}{\textbf{Favorita Dataset}} \\
\cline{2-4} 
\textbf{Functions} & \textbf{\textit{P10}}& \textbf{\textit{P50}}& \textbf{\textit{P90}} \\
\hline

Mean Absolute Error (MAE) & 0.642  & 0.590  & 0.691 \\ \hline
Mean Squared Error (MSE) & 0.472  & 0.447  & 0.420 \\ \hline
Mean Bias Error (MBE) & 1103.63  & 616.917  & 126.387 \\ \hline
\textbf{Relative Absolute Error (RAE)} & \textbf{0.463} & \textbf{0.376}  & \textbf{0.288} \\ \hline
Relative Squared Error (RSE) & 2.743  & 0.625  & 0.344 \\ \hline
Mean Absolute Percentage Error (MAPE) & 0.309  & 0.607  & 0.417 \\ \hline
Root Mean Squared Error (RMSE) & 0.519  & 0.462  & 0.404 \\ \hline
Mean Squared Logarithmic Error (MSLE) & 0.303  & 0.664  & 5.120 \\ \hline
Root Mean Squared Logarithmic Error (RMSLE) & 0.363  & 0.708  & 3.143 \\ \hline
Normalized Root Mean Squared Error (NRMSE) & 0.534  & 0.524  & 0.503 \\ \hline
Relative Root Mean Squared Error (RRMSE) &  0.662 & 0.552  & 0.511 \\ \hline
Huber Loss (delta = 0.5) &  0.574 &  0.584 & 0.522 \\ \hline
\textbf{Log Cosh Loss} & \textbf{0.336}  & \textbf{0.374}  & \textbf{0.396} \\ \hline
\textbf{Quantile Loss} &  \textbf{0.202} & \textbf{0.581}  & \textbf{0.274} \\ \hline
 
\end{tabular}
\label{tab1}
\end{center}
\end{small}
\end{table}

\begin{table}[h!]
\begin{small}
\caption{Summary of Loss Functions Performance on Volatility Dataset}
\begin{center}
\begin{tabular}{|p{4cm}|c|c|c|}
\hline
\textbf{Loss}&\multicolumn{3}{|c|}{\textbf{Volatility Dataset}} \\
\cline{2-4} 
\textbf{Functions} & \textbf{\textit{P10}}& \textbf{\textit{P50}}& \textbf{\textit{P90}} \\
\hline

\textbf{Mean Absolute Error (MAE)} & \textbf{0.042} & \textbf{0.041}  & \textbf{0.038} \\ \hline
Mean Squared Error (MSE) & 0.040  & 0.043  & 0.054 \\ \hline
Mean Bias Error (MBE) & 340.361  & 183.478  & 36.628 \\ \hline
Relative Absolute Error (RAE) & 0.037  & 0.042  & 0.054 \\ \hline
Relative Squared Error (RSE) & 0.149  & 0.140  & 0.043 \\ \hline
Mean Absolute Percentage Error (MAPE) & 0.123  & 0.074  & 0.023 \\ \hline
Root Mean Squared Error (RMSE) &  0.050  & 0.044  & 0.039 \\ \hline
Mean Squared Logarithmic Error (MSLE) & 0.020  & 0.121  & 0.276 \\ \hline
Root Mean Squared Logarithmic Error (RMSLE) &  0.040 & 0.332  & 0.530 \\ \hline
Normalized Root Mean Squared Error (NRMSE) & 7.952  & 2.425  & 3.772 \\ \hline
Relative Root Mean Squared Error (RRMSE) & 0.046  & 0.051  & 0.067 \\ \hline
\textbf{Huber Loss (delta = 0.5)} &  \textbf{0.037} & \textbf{0.040}  & \textbf{0.041} \\ \hline
Log Cosh Loss & 0.036  & 0.043  & 0.046 \\ \hline
\textbf{Quantile Loss} & \textbf{0.017} & \textbf{0.041}  & \textbf{0.020} \\ \hline
 
\end{tabular}
\label{tab1}
\end{center}
\end{small}
\end{table}

Table 16, 17, and 18 showcase the performance of loss functions on the electricity dataset, traffic dataset, and favorita dataset respectively.

Here are some of our observations:
\begin{itemize}
\item In case of Electricity dataset Quantile Loss, Mean Squared Error Loss, and Relative Root Mean Squared Error Loss have performed well.

\item In case of traffic dataset Log Cosh Loss, Quantile Loss, and Mean Absolute Error Loss have performed better.

\item In favorita dataset Relative Absolute Error Loss, LogCosh Loss, and Quantile Loss outperformed other objective functions.

\item In volatility dataset Quantile Loss, Mean Absolute Error Loss, and Huber Loss have performed better.

\end{itemize}

\section{Conclusion}

Loss functions play a critical role in determining a good fit model given the objective. Its' not possible to determine a universal loss function for complex objectives such as time series forecasting. There are a lot of factors such as outliers, skewness of data distribution, ML model requirements, computational requirements, and performance requirements. There is no single loss function that is applicable to all types of data. In academic setting, which focuses mostly on model architecture and data type, loss function can be determined by the data-set properties used for training, such as distribution, boundaries, etc. In this work, we have also attempted to structure the cases for which a particular loss function can be useful, such as in case of outliers in the datasets, Mean Squared Error is the best strategy; however if there are fewer outliers Mean Absolute Error would be a better choice than MSE. Similarly, Using LogCosh is the better approach if we want to keep it balance and if our objective is based on percentile Quantile loss would be a good choice.

In this study, we have summarized 14 well-known loss functions for time series forecasting and developed a tractable form of the loss function for improved and more accurate optimization. In future, we will use this work as a baseline implementation for time series forecasting applications.





\bibliographystyle{IEEEtran}
\nocite{*}
\bibliography{references}
\vspace{12pt}
\end{document}